\documentclass[preprint,12pt]{elsarticle}

\journal{你的目标期刊名称}

\usepackage{graphicx} 
\usepackage{subcaption}
\usepackage{amsmath}  
\usepackage{booktabs} 
\usepackage{multirow} 

\usepackage{ulem}
\usepackage{soul}
\sethlcolor{yellow}

\usepackage[table]{xcolor}
\definecolor{l0}{RGB}{255,255,255}
\definecolor{l1}{RGB}{255,230,183}
\definecolor{l2}{RGB}{255,208,111}
\definecolor{l3}{RGB}{247,170,88}
\definecolor{l4}{RGB}{239,138,71}
\definecolor{l5}{RGB}{231,98,84}
\definecolor{lightgreen}{RGB}{220, 255, 220}
\definecolor{lightblue}{RGB}{220, 240, 255}

\begin{document}

\begin{frontmatter}

\title{Dynamic Semantic Compression for Efficient Latent-Space Inference in Large Language Models}

\author{Peipei Li, Dongsen Zhang, Yuchen Liu, Wenjun Xu\corref{cor1}}

\ead{lipeipei@bupt.edu.cn, 
, liu20031225yuchen@bupt.edu.cn, wjxu@bupt.edu.cn}

\cortext[cor1]{Corresponding author}

\affiliation{organization={Beijing University of Posts and Telecommunications},
            city={Beijing},
            country={China}}

\begin{abstract}
Large Language Models (LLMs) primarily perform inference at the token level, resulting in substantial memory overhead and compromised computational efficiency. In this paper, we propose a Dynamic Semantic Extraction and Inference (DSEI) framework, which achieves segment-level inference within the latent space through a two-stage training strategy. First, we construct a Dynamic Semantic Autoencoder (DSAE) via self-supervised learning. DSAE dynamically extracts segment-level semantics and compresses them into compact latent representations via adaptive semantic weighting and gated fusion. Subsequently, we integrate the DSAE into the LLM architecture and train the model to infer over dense latent space. DSEI substantially reduces both input and generation sequences and significantly enhances inference efficiency. Extensive experiments conducted on the Wanjuan dataset demonstrate that DSEI reduces perplexity by 48\% compared to static sentence-level latent inference baseline. Furthermore, compared to standard LLMs using token-level inference, DSEI accelerates inference speed by 2.5× and reduces memory overhead by 90\%.
\end{abstract}


\end{frontmatter}

\section{Introduction}



\begin{figure*}[htbp]
    \centering
    \includegraphics[width=33pc,trim=1500 200 1600 200,clip]{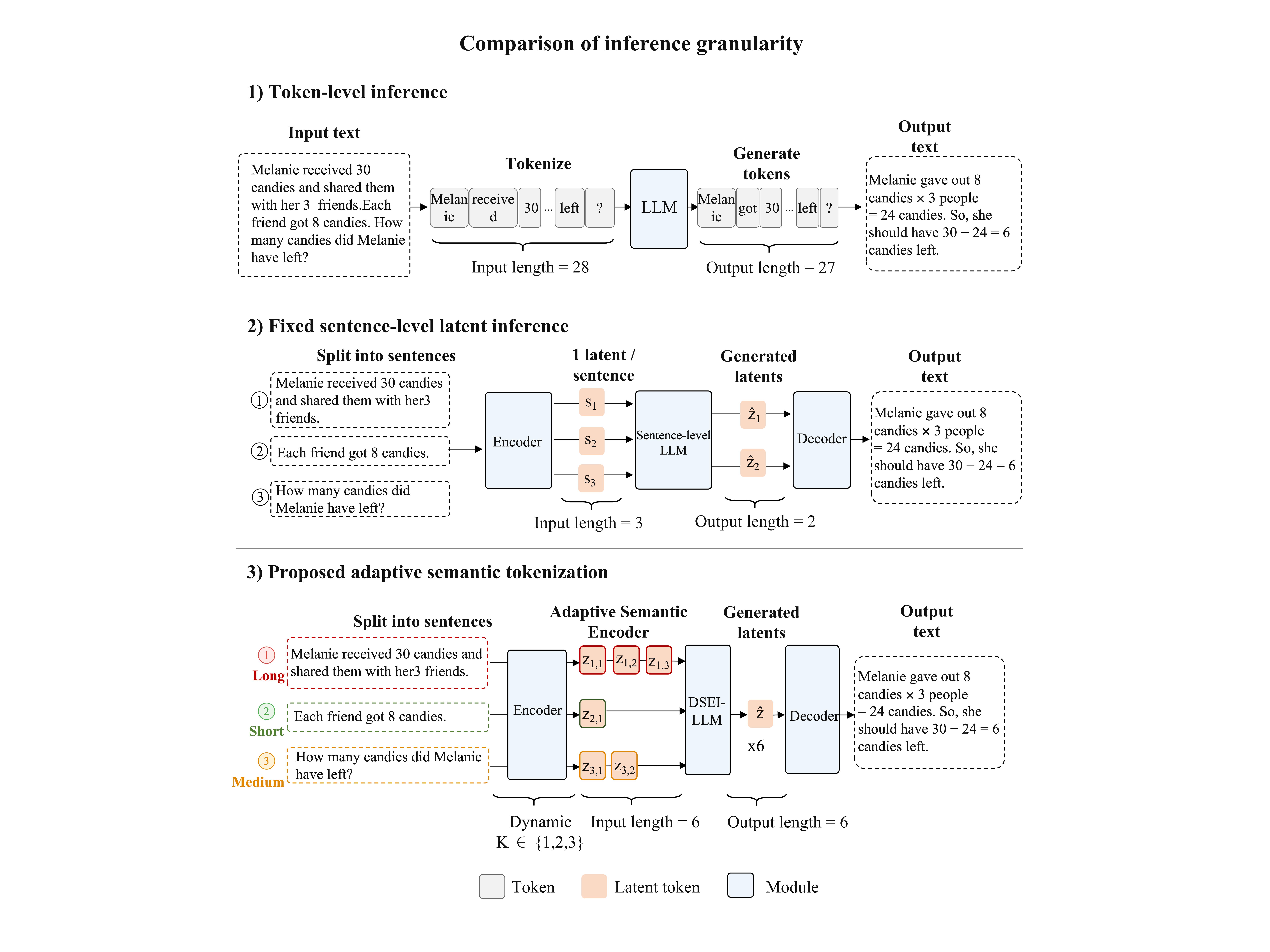}
    \caption{Comparison of token-level inference, fixed sentence-level latent inference, and the proposed dynamic segment-level latent inference. DSEI segments sentences according to token length and extracts compact semantic representations based on token-level importance, reducing both input and output sequence lengths while maintaining semantic fidelity.}
    \label{fig1}
\end{figure*}

{A}{utoregressive} large language models process and generate text at the token level. While this formulation provides fine-grained linguistic control, it also creates long sequential computation paths, intensive memory access, and high deployment costs as shown in Fig. \ref{fig1}(top). These costs become particularly restrictive when LLMs are deployed in resource-constrained or latency-sensitive engineering scenarios. Reducing the computational granularity of LLM inference from discrete tokens to compact semantic units is therefore a promising direction for efficient LLM deployment.

Recent latent-space inference methods take an important step in this direction by replacing token-level Chains-of-Thought (CoT) with latent representations, thereby reducing the number of autoregressive reasoning steps \cite{36}, \cite{37}, \cite{38}. While valuable, such approaches are generally limited to compressing the model’s intermediate reasoning process. In contrast, \cite{9} encodes each sentence into a latent representation, extending latent-space inference to the entire input-output stream as shown in Fig. \ref{fig1}(middle). However, this fixed granularity compression paradigm remains limited. It compresses every sentence into a single latent representation regardless of sentence length and treats all tokens uniformly during encoding. As a result, it overlooks both sentence-level and token-level differences in semantic density, which may lead to information loss or redundant compression and limit inference performance.


To address these limitations, we propose Dynamic Semantic Extraction and Inference (DSEI), a framework designed to improve the efficiency of LLM inference by shifting computation from token-level processing to adaptive segment-level latent-space inference as shown in Fig. \ref{fig1}(bottom). DSEI integrates a Dynamic Semantic Autoencoder (DSAE) into the LLM inference pipeline to extract compact yet informative semantic representations from text segments. Specifically, DSAE is first trained in a self-supervised manner to reconstruct input sentences from their latent representations. In the encoder, each sentence is divided into segments according to its token length, and token-level importance weights are estimated from hidden states to capture semantic contributions with different levels of information density. The weighted token representations are then aggregated to form dynamic semantic features, which are further fused with basic semantic features through a gating mechanism. The decoder is trained to reconstruct the original sentences from these representations. After this stage, the trained encoder and decoder are attached to the input and output sides of the LLM, respectively, and the overall model is further optimized through end-to-end training. In this way, DSEI reduces both input and output sequence lengths while maintaining semantic fidelity.

Extensive experiments on the Wanjuan-1.0 \cite{42} dataset demonstrate that, compared to static semantic extraction baselines, DSAE reduces perplexity by 34\% while maintaining computational efficiency. When integrated into LLMs, it achieves a 48\% reduction in perplexity (PPL). Furthermore, compared to LLMs that use token-level inference, DSEI reduces memory overhead by 90\% while simultaneously lowering PPL by 58\%.

Our main contributions are three-fold:

(1) We propose DSAE, a dynamic semantic compressor that partitions sentences into length-based semantic segments and extracts compact latent representations through token-level semantic weighting and gated fusion. This allows for efficient high-fidelity semantic extraction.

(2) We introduce the DSEI framework to integrate DSAE into LLMs, enabling end-to-end segment-level inference within the latent space, thereby reducing the effective sequence length of both input and output streams.

(3) Extensive experiments demonstrate that DSAE reduces perplexity by 34\% compared to static semantic extraction baselines, and by 48\% when integrated into LLM inference. Furthermore, compared to standard token-level LLM inference, DSEI reduces memory overhead by 90\% while lowering perplexity by 58\%, demonstrating a practical trade-off between generation quality and deployment efficiency.

\section{Related Work}

\subsection{Semantic compression}
Early research primarily focused on acquiring compact sentence vectors or instruction embeddings via contrastive learning \cite{1}, \cite{2}; these semantic representations are typically utilized for data retrieval and matching. Recent works have explored diverse semantic compression paradigms. For instance, GIST compresses prompts into a set of Transformer activations via meta-learning \cite{3}, while the LLMLingua model series \cite{5}, \cite{4} achieves hard compression by eliminating tokens with low PPL. Furthermore, ICAE demonstrates the feasibility of employing an encoder-decoder architecture for semantic compression \cite{6}. Building upon this trajectory, 500xCompressor \cite{7} enhances the compression ratio through KV cache compression, whereas C3 \cite{8} compresses semantics using cascaded small-parameter LLMs. SentenceVAE \cite{9}, on the other hand, focuses on sentence-level compression, encoding an entire sentence into a single sentence-level latent representation that can be reversibly reconstructed by a decoder. Another prominent direction involves continuous latent space modeling, which achieves semantic compression via diffusion models \cite{10}, \cite{11}, \cite{12}, \cite{13}. Within this domain, AR-Diffusion \cite{14} improves parallelism and coherence through local causal constraints, while two-stage methods such as LD4LG \cite{15} and PLANNER \cite{16} enhance controllability and efficiency via a compression-then-refinement approach. Unlike fixed sentence-level compression, DSEI uses length-adaptive segmentation and token-level semantic weighting to avoid over-compressing long sentences.

\subsection{Latent-space inference}
Latent reasoning methods aim to perform inference within continuous latent spaces and can be broadly categorized into three primary directions: knowledge internalization, architectural modification, and autoregressive latent reasoning.
Knowledge internalization seeks to directly embed logic and reasoning capabilities into model parameters, thereby reducing reliance on explicit chains of thought or explanatory prompts. For instance, iCoT-SI \cite{17} forces the model to internalize reasoning structures by progressively eliminating explicit reasoning steps during training, whereas TwT \cite{18} emphasizes the roles of multi-teacher distillation and habitual reasoning. Pause \cite{19} encodes reasoning steps into dedicated token embeddings, while CoCoMix \cite{20} demonstrates that reasoning capabilities can be implanted through continuous concept mixing during the pre-training phase. Furthermore, \cite{21} reveals the critical impacts of reasoning granularity, representation format, and teacher model selection on distillation efficiency.
Architectural modification leverages the hierarchical structure of Transformers to enable selective reasoning and dynamic computation \cite{22}, \cite{23}, \cite{24}, \cite{25}, \cite{26}. Approaches like \cite{28} allow models to dynamically adjust computational steps according to problem difficulty, whereas \cite{29}, \cite{30} explore supervision and guidance oriented toward implicit trajectories. Other studies  \cite{31}, \cite{32}, \cite{34}, \cite{35} focus on novel architectures and parallel inference.
Autoregressive latent reasoning utilizes hidden states to represent the thought process \cite{36}, \cite{37}. In this vein, CODI \cite{38} constructs an autoregressive latent variable model combined with self-distillation, while \cite{39}, \cite{40} concentrate on enhancing the stability of these hidden states.
In contrast to latent inference methods that primarily compress intermediate reasoning trajectories into continuous representations, DSEI extends latent-space inference to the complete input-output stream through segment-level semantic representations.

\section{Method}

\begin{figure*}[htbp]
    \centering
    \includegraphics[width=33pc,trim=0 1800 0 1200,clip]{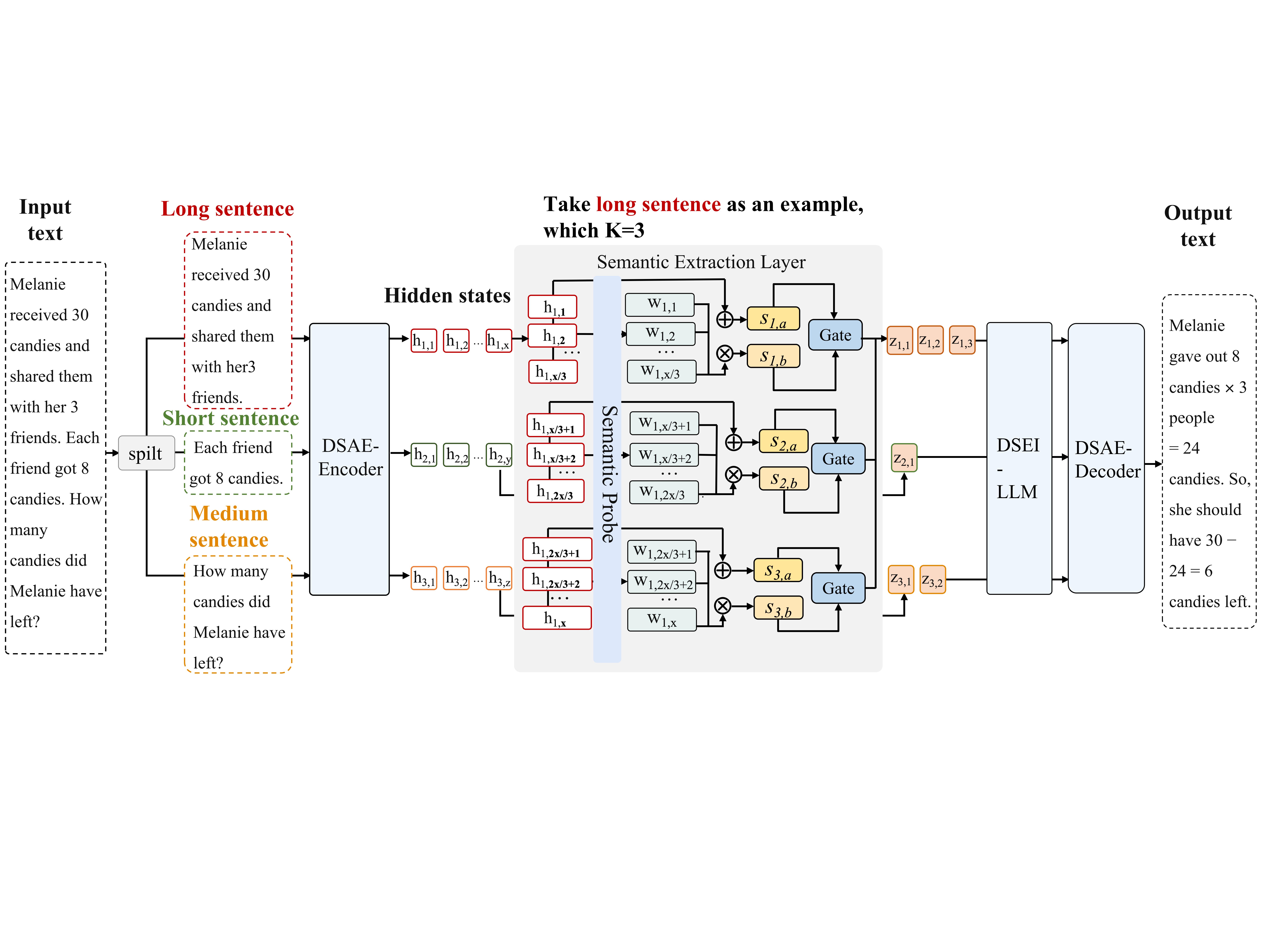}
    \caption{Overall architecture of the proposed DSEI framework. The input text is first split into sentences, and each sentence is partitioned into one or more segments according to token length. The DSAE encoder extracts segment-level latent representations through token-level semantic weighting and gated fusion. These latent representations are then processed by the DSEI-LLM for latent-space inference, and the generated latent outputs are reconstructed into natural language by the DSAE decoder.}
    \label{fig2}
\end{figure*}

In this section, we introduce the proposed DSEI. Given a vanilla LLM $\mathcal{L}$ and an input text consisting of $n$ sentences $\mathbf{x} =\left \{ x^1,x^2,...,x^n \right \} $, the text is tokenized and embedded to obtain $m$ token embedding vectors $\mathbf{e} =\left \{ e_{1:l_1}^1,e_{1:l_2}^2,...,e_{1:l_n}^n \right \} $, where $e_{1:l_n}^n$ denotes the $l_n$ embedding vectors of the $n$-th sentence $x^n$, and $m=\sum_{i=1}^{n}l_i$. $\mathcal{L}$ processes these embeddings and predicts the next token, generating $j$ sentences comprising $t$ tokens $\mathbf{o} =\left \{ o_{1:b_1}^1,o_{1:b_2}^2,...,o_{1:b_j}^j \right \} $, where $o_{1:b_j}^j$ represents the $b_j$ tokens of the $j$-th sentence, and $t=\sum_{i=1}^{j}b_i$.

To address the issue of excessively long inputs and outputs in LLMs, we propose enabling the LLM to perform inference within a segment-level latent space, thereby reducing the input length from $m$ to $\sum_{i=1}^{n}K_i$, where each sentence is partitioned into $K_i$ segments. Similarly, the output length is reduced from $t$ to $\sum_{r=1}^{j}\widehat{K}_r$, where $\widehat{K}_r$ denotes the number of latent segments associated with the r-th output sentence. This requires our method to: i) extract the semantics of sentences and efficiently compress them into the latent space; ii) enable the LLM to comprehend segment-level latent representations and perform segment-by-segment inference. DSEI is designed precisely to achieve these two objectives.

\subsection{Semantic Autoencoder}
\label{SAE}
As shown in Fig. \ref{fig2}, the input to DSAE is a single sentence, which can be represented as $e_{1:l}^1$ after tokenization and embedding.
The encoder is composed of stacking Transformer encoder layers \cite{53},
which map these embedding vectors into $l$ hidden states $\mathbf{h}=h_{1:l} $. These hidden states $\mathbf{h}$ are partitioned  into $K$ segments according to the token length. Specifically, sentences with fewer than 15 tokens are treated as a single segment, those with 15--30 tokens are partitioned into two, and longer sentences are partitioned into three segments. Formally, the number of segments is determined by the token length $l$:

\begin{equation}
\displaystyle
\mathbf{h}^{k} = \left\{ h_i \;\middle|\; \left\lfloor \frac{k \cdot l}{K} \right\rfloor \le i < \left\lfloor \frac{(k+1) \cdot l}{K} \right\rfloor \right\}, \quad k = 0, 1, \dots, K-1,
\end{equation}

\noindent where $\mathbf{h}^{k}$ denotes the hidden states belonging to the $k$-th segment.

To extract the segment semantics based on $\mathbf{h}^{k}$, a straightforward approach is to sum $\mathbf{h}^{k}$, as shown in Eq. \ref{eq:static}. Although the resulting latent representation $s_a$ retains global information, compressing hidden states containing varying amounts of information equally into one vector also introduces redundancy. To address this issue, we employ a fully connected layer called the Semantic
Probe to map $\mathbf{h}^{k}$ to one dimension, obtaining a sequence of token weights $\mathbf{w}^{k}$. A softmax function is applied to normalize
these weights into attention probabilities. Since $\mathbf{h}^{k}$ obtained by the
self-attention mechanism is context-dependent, $\mathbf{w}^{k}$ can dynamically
identify the semantic informativeness of tokens. The latent representation
$s_b$ is obtained by the weighted average of
$\mathbf{w}^{k}$ and $\mathbf{h}^{k}$, as shown in Eq. \ref{eq:dynamic}:

\begin{equation}
\label{eq:static}
s_{k,a}=\mathrm{LayerNorm}\left(\sum \mathbf{h}^{k} \right)
\end{equation}

\begin{equation}
\label{eq:dynamic}
s_{k,b}=\mathrm{LayerNorm}\left(\mathrm{softmax}(\mathrm{linear}(\mathbf{h}^{k}))\mathbf{h}^{k\top}\right)
\end{equation}

For the reconstruction task of the decoder, the signal provided by the dynamically extracted $s_{k,b}$ is too sparse, making it difficult for the decoder to predict a reconstructed output consistent with the input. To address this issue, we combine $s_{k,a}$, which contains global information, with $s_b$ through a learnable scalar $w_{gate}$. This scalar is mapped to a value between 0 and 1 via a sigmoid function, implementing a simple yet effective gating mechanism that balances the contribution of the two representations. The resulting segment latent representation $s_k$ thus incorporates global information while reducing redundancy. This process can be formally represented as:

\begin{equation}
    s_k=\mathrm{sigmoid}(w_{gate})\cdot s_{k,a}+(1-\mathrm{sigmoid}(w_{gate}))\cdot s_{k,b}
\end{equation}


Given a sentence represented by a sequence of segment-level latent representations $\mathbf{s}=s_{1:K}$, the decoder reconstructs the original sentence from all segment representations jointly. The decoder consists of stacked Transformer decoder layers, where $\mathbf{s}$ is used as the key and value in cross-attention. Starting from the start token, the decoder autoregressively generates the reconstructed token probabilities $\mathbf{p}=p_{1:y}$, where $p_y$ denotes the predicted probability of the y-th target token. The reconstruction probabilities are then compared with the original input tokens using focal loss \cite{45}, which serves as the training objective in the first stage. This process can be formulated as:

\begin{equation}
    \mathcal{L}_{\mathrm{comp}} =\sum_{i=1}^{y} \mathrm{FL}(p_i)=\sum_{i=1}^{y}-(1-p_i)^2\mathrm{log}(p_i) 
\end{equation}

The initial parameters of the encoder and decoder in DSAE are transferred from the LLM backbone of DSEI. The encoder and decoder in DSAE maintain the same number of layers. For a DSAE with $i$ layers, since the encoder is responsible for extracting features from text embeddings, its parameters are transferred from the first $i$ layers of the functionally equivalent LLM. Similarly, the decoder is responsible for generating text from latent representations, and its parameters are transferred from the last $i$ layers of the functionally equivalent LLM. Compared to random initialization, parameter transfer endows the model with semantic comprehension capabilities, allowing training to proceed without starting from basic grammatical structure construction. This enables efficient training of the model's semantic extraction abilities, leading to convergence at better performance.

\subsection{Segment-by-segment inference}

To enable the LLM to perform segment-by-segment inference in the latent space, we integrate DSAE into the LLM and optimize the resulting model in an end-to-end manner. In DSEI, the token embedding layer of the vanilla LLM is replaced by the DSAE encoder and the semantic extraction layer, while the output hidden states of the LLM are decoded into natural language by the DSAE decoder. Given an input text, we first split it into $n$ sentences by pattern matching based on punctuation marks. For the $i$-th sentence $x^{i}$, the DSAE encoder and semantic extraction layer determine the number of segments $K_i$ according to its token length and encode it into a sequence of segment-level latent representations $s^{i}_{1:k_{i}}$. The latent representations of all sentences are then concatenated in their original textual order to form the LLM input $\mathbf{s}=\left \{{s^{1}_{1:k_{1}},s^{2}_{1:k_{2}},...,s^{n}_{1:k_{n}}}\right \}$, where the total number of latent vectors is $\sum_{i=1}^{n}K_i$. This encoding process is performed in parallel across sentences in a batched manner. The LLM takes $\mathbf{s}$ as input and autoregressively generates a sequence of output latent representations $\widehat{\mathbf{s}}=\widehat{s}_{1:T}$.

Since the LLM produces a flat sequence of segment-level latent representations, DSEI further introduces a segmentation head to recover sentence-level structure before decoding. Specifically, for each generated latent representation $\widehat{s}_t$, the segmentation head predicts whether it marks the end of a sentence-level latent group. These predicted boundaries partition $\widehat{\mathbf{s}}$ into several groups, each corresponding to one output sentence. The DSAE decoder then reconstructs a natural-language sentence from each latent group, and the concatenation of all reconstructed sentences forms the final output of DSEI.

To determine when the LLM should stop generating segment-level latent
representations, we train a fully connected layer called the termination head. During training, the termination objective is computed only for generated latent representations that correspond to ground-truth sentence-level latent group boundaries. Given such a boundary representation $\widehat{s}_i$, the termination head maps it to a
two-dimensional logit vector $\mathbf{d}_{i}=[d_i^1,d_i^2]$, where the two dimensions indicate whether generation should continue or terminate. If the
vector signals an end state, the generation iteration terminates; otherwise,
the LLM continues to generate the next latent representation.

During training, we use focal loss \cite{45} to optimize the segmentation head, the termination head, and the latent-generation process jointly. Formally, the total loss is defined as:

\begin{equation}
    \displaystyle
    \mathcal{L}_{\mathrm{infer}} =
    \underbrace{\frac{\lambda}{j}\sum_{i=1}^{t} \mathrm{FL}(p_i)}_{\mathrm{generation\;term}} +
    \underbrace{\sum_{i=1}^{t} \mathrm{FL}({b}_{i})}_{\mathrm{segment\;term}}+
    \underbrace{\sum_{t\in \mathcal{B}} \mathrm{FL}(d_i)}_{\mathrm{stop\;term}}
    ,
\end{equation}

where $\lambda$ is a weighting coefficient used to balance the token-level generation term with the stop and segment terms, $b_t$ denote the predicted probability assigned to the ground-truth boundary label of $\widehat{s}_t$, $d_i$ denote the predicted probability assigned to the ground-truth termination label of $\widehat{s}_t$, and $\mathcal{B}$ is the set of boundary positions at the end of sentence.

\section{Experiments}

We first train DSAE via self-supervised learning, and subsequently integrate the trained DSAE into the LLM for joint end-to-end training. We analyze DSAE and DSEI separately, comparing them with baseline methods and token-by-token LLMs, exploring the characteristics of dynamic semantics, and analyzing the contributions of different components.

\subsection{Experimental setup}

\textbf{Dataset and metrics.} For both DSAE and DSEI, all experiments are trained and validated on the English subset of the Wanjuan-1.0 dataset \cite{42}. For DSAE, the training set samples approximately 155M sentences, and the validation set consists of 1K non-overlapping sentences. For DSEI, the training set samples approximately 6.4M paragraphs, and the validation set consists of 1K non-overlapping paragraphs. Following \cite{9}, we use PPL as the evaluation metric for DSAE and DSEI. Additionally, we employ mean input throughput and mean GPU memory to assess the computational efficiency of DSEI.

\textbf{Baseline methods.} We compare our method with the baseline approach \cite{9}, which includes SVAE, an autoencoder with a static fusion mechanism for sentence compression and reconstruction, and SLLM, a sentence-level LLM that integrates SVAE. In SVAE, the sentence-level latent representation is obtained by summing the hidden states from the last layer of the encoder, and SLLM performs sentence-level inference based on this static latent representation. Additionally, we also compare with token-level inference LLMs from the OPT \cite{44} series.

\textbf{Implementation details.} (1) Base Models: We adopt three OPT \cite{44} series models of sizes 125M, 350M, and 1.3B as our base LLMs to validate the effectiveness of our method across different model scales. The dimension of hidden states in DSAE is kept consistent with that of the base LLM. During weight transfer, both the encoder and decoder only utilize the weights of the self-attention modules from the base LLM, excluding the feed-forward network (FFN) weights. (2) Hyperparameters: For DSAE, the maximum input token length is set to 64, batch size to 512, learning rate to 1e-7, and the gating scalar $w_{gate}$ is initialized to 0. In experiments where the base LLM is 125M, the number of layers in DSAE is set to 1, 2, and 4 respectively, while for base LLMs of other sizes it is set to 1. For DSEI, the maximum input sentence length is set to 64, batch size is 4, and the learning rate is 1e-6. For all experiments, $\lambda$ is set to 0.01. We use the AdamW optimizer \cite{43} with a weight decay of 1e-2 and gradient clipping with a maximum L2 norm of 1. The learning rate follows a linear schedule for the first 5,000 iterations, and then uses a cosine annealing schedule for subsequent iterations. All experiments are conducted on a single RTX 5880 Ada Generation GPU.

\subsection{Semantic extraction results}

\begin{figure}[tbp] %
    \centering   
    
    \begin{subfigure}[b]{0.45\textwidth} 
        \centering
        \includegraphics[scale=0.3]{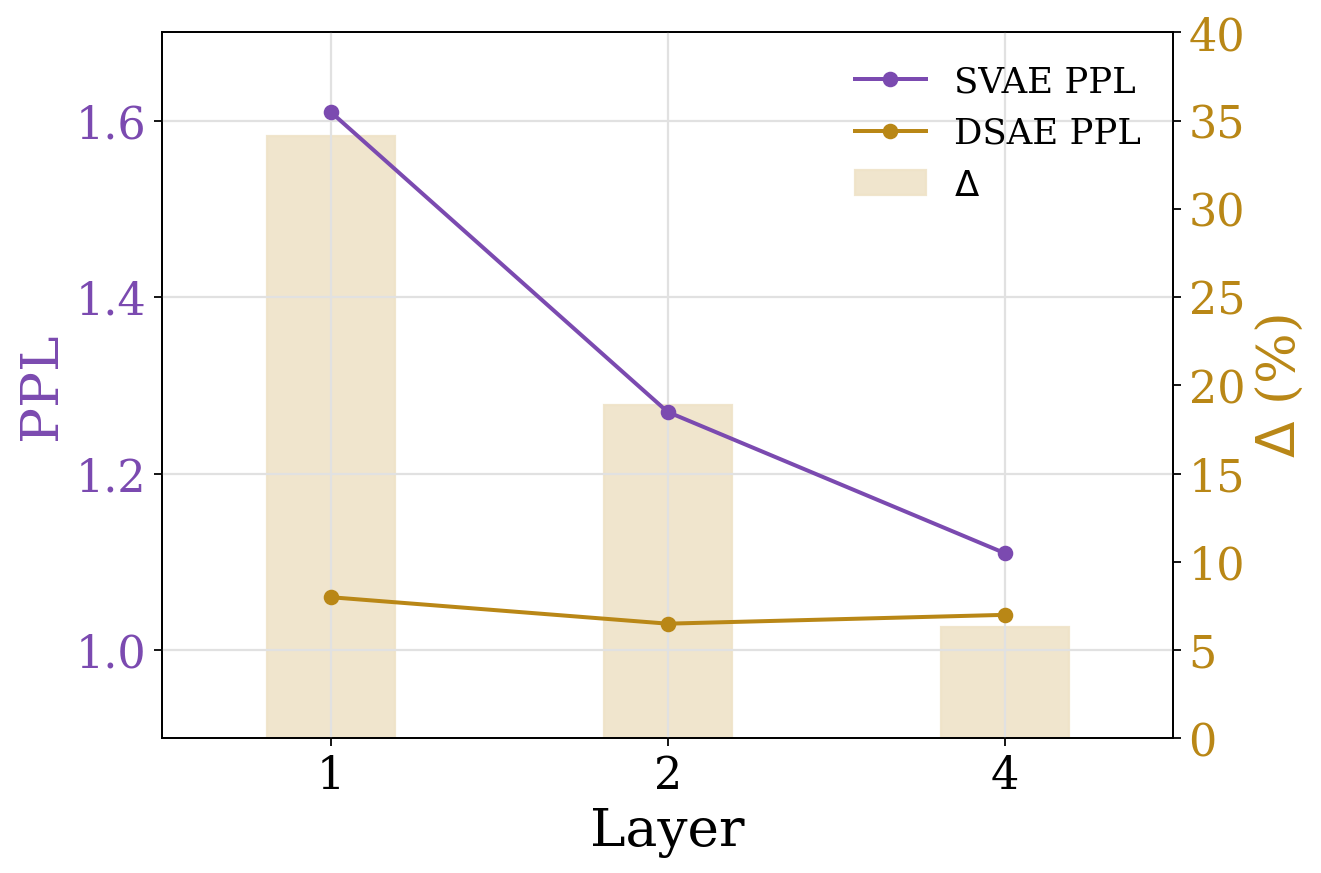}
        \caption{PPL for different layers with 768 hidden size.} 
    \end{subfigure}
    \hfill 
    \begin{subfigure}[b]{0.45\textwidth} 
        \centering
        \includegraphics[scale=0.3]{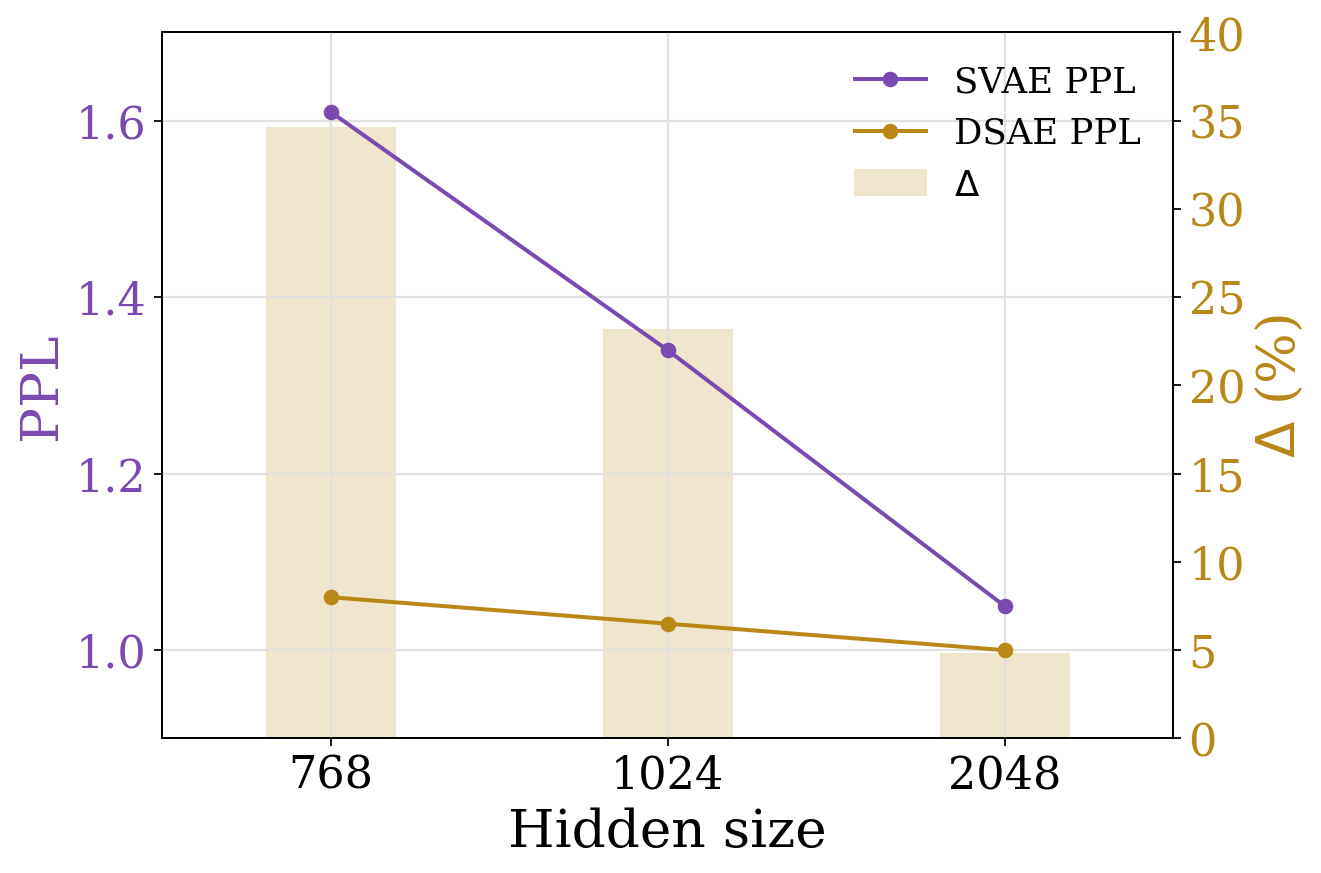}
        \caption{PPL for different hidden sizes with 1 layer.} 
    \end{subfigure}
    
    \caption{Comparison of PPL for semantic extraction between DSAE and baseline method across varying (a) layers and (b) hidden sizes. $\bigtriangleup$ denotes percentage decrease in PPL of DSAE compared to the baseline method.}
    \label{fig3}
\end{figure}

Fig. \ref{fig3} compares the performance of DSAE with the baseline method SVAE under different model sizes. DSAE demonstrates consistent performance improvements over the static representation fusion-based semantic extraction method across various model depths (varying layers) and widths (varying hidden sizes). Notably, in the smallest model configuration, DSAE achieves a 34\% reduction in PPL compared to the baseline method, which is the largest performance gain among all model sizes. This is particularly advantageous for resource-constrained edge devices, which typically employ smaller-scale models.

\begin{table*}[]
\centering
\caption{Test examples of sentence reconstruction between baseline method and DSAE. The darker the color, the greater the weight of the token. Text in red denotes mismatches between reconstructed sentence and input sentence, and \textcolor{red}{$\triangle$} denotes missing output.}
\label{tab:weight and rebuild}
\resizebox{\textwidth}{!}{%
\begin{tabular}{lll}
\toprule[1.2pt]
\multicolumn{1}{c}{\multirow{2}{*}{\textbf{Weight of input sentence}}}                                                     & \multicolumn{2}{c}{\textbf{Output sentence}}                                                                                                                                                                                                     \\ \cmidrule(r){2-3}
\multicolumn{1}{c}{}                                                                                                       & \multicolumn{1}{c}{\textbf{SVAE}}                                                                                   & \multicolumn{1}{c}{\textbf{DSAE}}                                                                                          \\ \midrule

\begin{tabular}[c]{@{}l@{}}
\colorbox{l3}{\strut Some} \colorbox{l3}{\strut also} \colorbox{l3}{\strut have} \colorbox{l3}{\strut performance} \colorbox{l3}{\strut venues} \colorbox{l1}{\strut for} \colorbox{l4}{\strut various} \\[5pt] 
\colorbox{l3}{\strut kinds} \colorbox{l2}{\strut of} \colorbox{l3}{\strut artists} \colorbox{l2}{\strut .} 
\end{tabular}                 
& \begin{tabular}[c]{@{}l@{}}Some also have performance venues for various \\[5pt] kinds of artists.\end{tabular}          
& \begin{tabular}[c]{@{}l@{}}Some also have performance venues for various \\[5pt] kinds of artists.\end{tabular}                 \\ \midrule

\begin{tabular}[c]{@{}l@{}}
\colorbox{l3}{\strut Intense} \colorbox{l4}{\strut pulsed} \colorbox{l2}{\strut light} \colorbox{l2}{\strut device} \colorbox{l2}{\strut emit} \colorbox{l1}{\strut a} \colorbox{l2}{\strut range} \\[5pt] 
\colorbox{l2}{\strut wavelength} \colorbox{l2}{\strut (} \colorbox{l3}{\strut 515} \colorbox{l1}{\strut -} \colorbox{l2}{\strut 1200} \colorbox{l2}{\strut )} \colorbox{l1}{\strut of} \colorbox{l2}{\strut light} \colorbox{l1}{\strut .} 
\end{tabular}         
& \begin{tabular}[c]{@{}l@{}}Intense pulsed light device emit a range \textcolor{red}{of} \\[5pt] \textcolor{red}{puls}(\textcolor{red}{660}-1200) \textcolor{red}{$\triangle$} light.\end{tabular}        
& \begin{tabular}[c]{@{}l@{}}Intense pulsed light device emit a range \\[5pt] wavelength(515-1200) of light.\end{tabular}         \\ \midrule

\begin{tabular}[c]{@{}l@{}}
\colorbox{l2}{\strut Makes} \colorbox{l1}{\strut the} \colorbox{l2}{\strut perfect} \colorbox{l2}{\strut gift} \colorbox{l1}{\strut for} \colorbox{l5}{\strut Schylling} \colorbox{l2}{\strut enthusiasts} \\[5pt] 
\colorbox{l1}{\strut that} \colorbox{l2}{\strut are} \colorbox{l1}{\strut at} \colorbox{l2}{\strut least} \colorbox{l2}{\strut 3} \colorbox{l2}{\strut years} \colorbox{l2}{\strut old} \colorbox{l1}{\strut .} 
\end{tabular} 
& \begin{tabular}[c]{@{}l@{}}Makes the perfect gift for \textcolor{red}{Schreferably that} \\[5pt] \textcolor{red}{enthusiasts} at \textcolor{red}{$\triangle$} 3 years old.\end{tabular} 
& \begin{tabular}[c]{@{}l@{}}Makes the perfect gift for Schylling enthusiasts \\[5pt] that are at least 3 years old.\end{tabular} \\ \midrule

\colorbox{l3}{\strut His} \colorbox{l3}{\strut hat} \colorbox{l2}{\strut is} \colorbox{l5}{\strut accented} \colorbox{l1}{\strut with} \colorbox{l5}{\strut holly} \colorbox{l1}{\strut and} \colorbox{l5}{\strut holly} \colorbox{l3}{\strut berries} \colorbox{l1}{\strut .} 
& His hat is accented with holly \textcolor{red}{holly} and berries.                                                                   
& His \textcolor{red}{holly} accented with holly and holly berries.                                                                                                                                                   \\ \bottomrule[1.2pt]
\end{tabular}%
}
\label{tab2}
\end{table*}

\begin{table}[ht]
\centering
\caption{The ablation experiment results of DSAE. $\bigtriangleup$ denotes percentage increase in PPL compared to the default setting. ``First layer weights'' indicates that the initial parameters for both the encoder and the decoder are transferred from the first layer of the base LLM, while ``Last layer weights'' are derived from the final layer.}
\label{tab:ablation}
\resizebox{0.5\columnwidth}{!}{%
\begin{tabular}{lll}
\toprule[1.2pt]
\textbf{Method}     & \textbf{PPL$\downarrow$} & \textbf{$\bigtriangleup$}   \\ \midrule
Default             & 1.06         & 0.00\%         \\
w/o Segmentation    & 1.40         & 32.07\%         \\
w/o $s_a$            & 1.14         & 7.55\%          \\
w/o $s_b$            & 1.16         & 9.43\%          \\
w/o gate            & 1.21         & 14.15\%          \\
w/o LLM weights     & 1.24         & 16.98\%         \\
First layer weights & 1.13         & 6.60\%          \\
Last layer weights  & 1.17         & 10.38\% \\ \bottomrule[1.2pt]
\end{tabular}%
}
\label{tab3}
\end{table}

Tab. \ref{tab2} presents the sentence reconstruction performance of DSAE compared with the baseline method across multiple test cases. DSAE demonstrates more accurate sentence reconstruction compared to the baseline method. The token weight results for input sentences show that DSAE pays greater attention to content words rich in semantic information, such as ``various'' in sample 1 and ``pulsed'' in sample 2, while assigning lower weights to function words that lack substantial meaning, such as ``for'', ``a'', and ``of''. This indicates that DSAE can efficiently identify key information in sentences based on context and compress it into a compact latent representation, exhibiting particular advantages when processing longer sentences or those containing uncommon words. For instance, in sample 2, SVAE reconstructed ``515'' as ``660'', while DSAE assigned greater weight to this token during semantic extraction and successfully reconstructed the original token. In sample 3, which contains the proper noun ``Schylling'', DSAE assigned it the highest weight and accurately generated the word.

To validate the effectiveness of each component in DSAE, we conducted ablation studies on a model with hidden size 768 and 1 layer, with results shown in Tab. \ref{tab3}. Four key findings emerged:

(1) Among all ablated components, sentence segmentation yields the most pronounced effect on semantic compression. Without segmentation, PPL increases by 32\% relative to the default setting. This demonstrates that forcing an entire sentence into a single latent representation creates a severe information bottleneck. In contrast, the proposed segmentation strategy assigns multiple latent vectors to longer sentences according to their token length, allowing the compressed representation to preserve richer semantic details.

(2) Static and dynamic latent representations provide complementary semantic information. We directly used the static latent representation $s_a$ and dynamic latent representation $s_b$ separately as sentence semantics for the decoder to reconstruct sentences. We found that when the static latent representation is absent, reconstructed sentences tend to miss words, while when the dynamic latent representation is absent, inconsistent words are reconstructed—both leading to performance degradation.

(3) The gating fusion mechanism is beneficial. In the gating ablation experiment, we simply summed the static latent representation $s_a$ and dynamic latent representation $s_b$ as the sentence latent representation. this simple summation strategy increases PPL by 14\%. This confirms the importance of gated fusion of global and dynamic semantic information when extracting sentence semantics.

(4) LLM pre-trained parameters are suitable for semantic extraction. When initializing model parameters randomly instead of transferring from the LLM, performance dropped by 17\%. This indicates that the language understanding capabilities acquired from LLM pre-training are well-suited for semantic extraction tasks. Additionally, we investigated the suitability of different LLM layers. When initializing encoder and decoder parameters with the first and last layers of the LLM respectively, performance decreased by 6\% and 10\%. This suggests that shallow layers align better with the encoder—both are responsible for extracting features from text—while deep layers align better with the decoder—both are responsible for generating text based on extracted features.

\begin{figure}[htbp]
    \centering
    \begin{subfigure}[b]{1\textwidth}
        \centering
        \includegraphics[width=\textwidth,height=5cm]{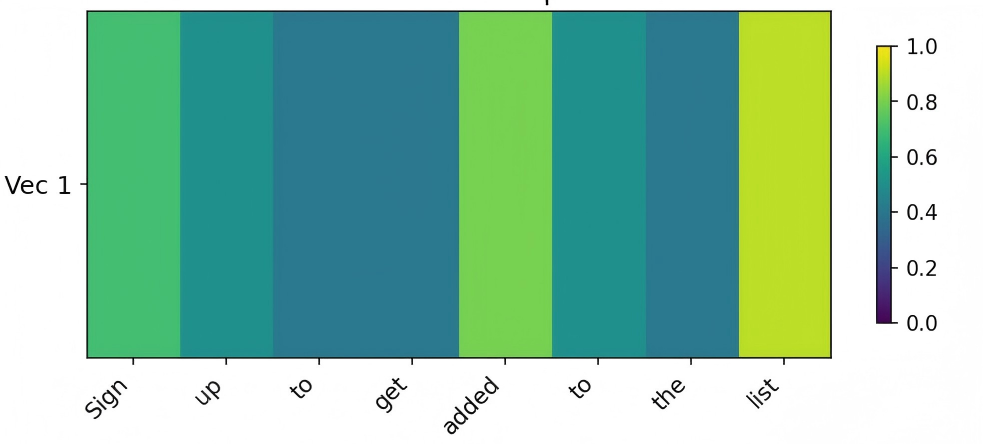}
        \caption{Short sentence.}
    \end{subfigure}
    \vspace{0.3em}
    \begin{subfigure}[b]{1\textwidth}
        \centering
        \includegraphics[width=\textwidth]{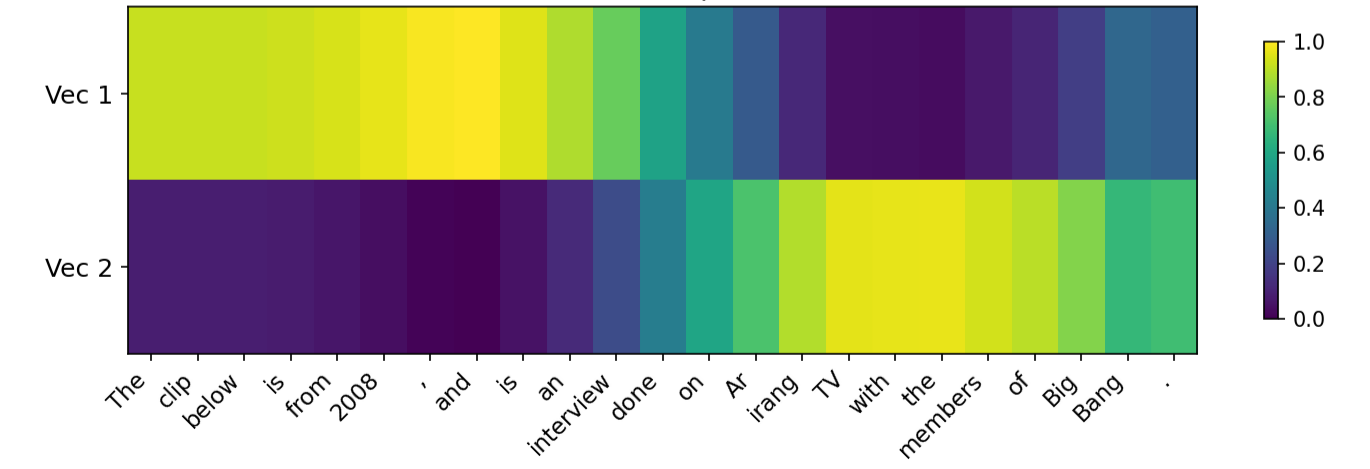}
        \caption{Middle sentence.}
    \end{subfigure}
    \vspace{0.3em}
    \begin{subfigure}[b]{1.07\textwidth}
        \centering
        \includegraphics[width=\textwidth]{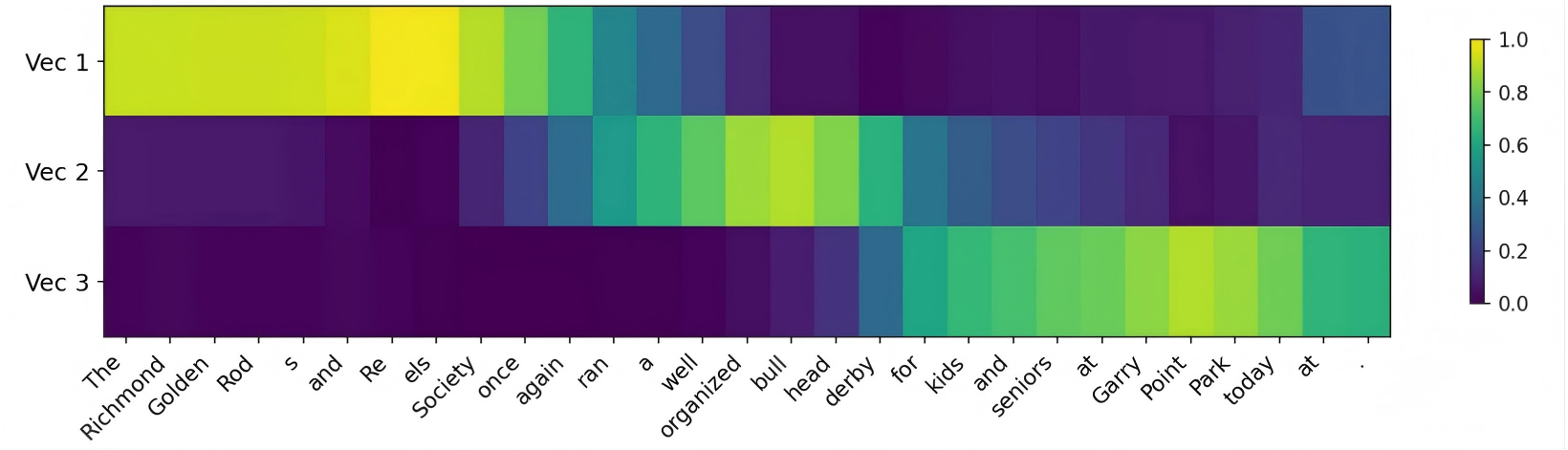}
        \caption{Long sentence.}
    \end{subfigure}
    \caption{Attention heatmap comparison of decoder cross-attention weights on DSAE tokens for (a) short, (b) middle, and (c) long sentences during semantic compression. Lighter colors indicate higher attention weights, signifying that the decoder assigns greater importance to those tokens when reconstructing the original sentence.}
    \label{fig:heatmap}
\end{figure}

To further analyze how the decoder utilizes segment-level latent representations, we visualize the cross-attention weights between latent vectors and tokens for sentences of different lengths, as shown in Fig.~\ref{fig:heatmap}. For short sentences, the single latent vector exhibits a broad attention pattern over the entire sentence, indicating that one compact representation is sufficient for reconstruction when the semantic content is compact. For medium and long sentences, different latent vectors exhibit clear and complementary attention patterns over different token regions. This suggests that length-based segmentation enables multiple latent vectors to capture localized semantic information, thereby reducing the burden on a single sentence-level vector and improving the fidelity of sentence reconstruction, thus pointing to a more effective design for long-text semantic compression.

\subsection{Latent inference results}

\begin{table*}[]
\centering
\caption{Experiment results of baseline methods and DSEI. DSEI-b-l denotes an OPT model with parameter size b, integrated with a DSAE consisting of l layers.}
\label{tab:llm}
\resizebox{\textwidth}{!}{%
\begin{tabular}{ccccc}
\toprule[1.2pt]
\textbf{Model} & \textbf{Parameters(M)} & \textbf{PPL$\downarrow$} & \textbf{\begin{tabular}[c]{@{}c@{}}Mean input throughput\\ (k tokens/s)$\uparrow$\end{tabular}} & \textbf{\begin{tabular}[c]{@{}c@{}}Mean GPU memory\\ (KB/token)$\downarrow$\end{tabular}} \\ \midrule
OPT-125M       & 125.23                 & 26.94        & 13.93                                                                                 & 131.45                                                                        \\
\rowcolor[HTML]{EFEFEF} 
SLLM-125M-H1   & 214.32                 & 21.84        & 35.90                                                                                 & 10.97                                                                         \\
\rowcolor[HTML]{EFEFEF} 
DSEI-125M-H1   & 214.92                 & 11.17        & 35.98                                                                                 & 13.05                                                                         \\
SLLM-125M-H2   & 226.14                 & 21.20        & 37.49                                                                                 & 9.94                                                                          \\
DSEI-125M-H2   & 226.76                 & 15.52        & 37.76                                                                                 & 13.18                                                                         \\
\rowcolor[HTML]{EFEFEF} 
SLLM-125M-H4   & 249.78                 & 20.99        & 24.36                                                                                 & 10.60                                                                         \\
\rowcolor[HTML]{EFEFEF} 
DSEI-125M-H4   & 250.37                 & 14.57        & 23.50                                                                                 & 13.56                                                                         \\ \midrule
OPT-350M       & 331.19                 & 22.60        & 7.71                                                                                  & 184.72                                                                        \\
\rowcolor[HTML]{EFEFEF} 
SLLM-350M-H1   & 429.46                 & 21.51        & 99.20                                                                                 & 25.21                                                                         \\
\rowcolor[HTML]{EFEFEF} 
DSEI-350M-H1   & 432.66                 & 14.12        & 98.30                                                                                & 27.85                                                                         \\ \midrule
OPT-1.3B       & 1315.75                & 16.02        & 7.9                                                                                   & 285.24                                                                        \\
\rowcolor[HTML]{EFEFEF} 
SLLM-1.3B-H1   & 1605.67                & 19.30        & 60.88                                                                                 & 50.44                                                                         \\
\rowcolor[HTML]{EFEFEF} 
DSEI-1.3B-H1   & 1609.88                & 9.86        & 59.53                                                                                 & 52.07                                                                         \\ \bottomrule[1.2pt]
\end{tabular}%
}
\end{table*}

Tab. \ref{tab:llm} presents a comparison between DSEI and existing baseline methods in terms of PPL, throughput, and GPU memory overhead. Experimental results demonstrate that DSEI achieves an optimal balance between inference efficiency and accuracy. Across various parameter scales of LLM backbones, DSEI consistently exhibits improved PPL performance compared to static sentence-level latent inference baseline SLLM.

It is worth noting that the quality improvement brought by DSEI is particularly pronounced on smaller-scale LLMs, while its throughput remains comparable to SLLM and its memory overhead remains substantially lower than token-level inference. On a 125M‑parameter model, compared with SLLM, a single‑layer DSEI configuration achieves a 48\% reduction in PPL, although the average input throughput increases only marginally by 0.22\% and the mean GPU memory overhead increases by 18\%. These observations suggest that DSEI may be particularly suitable for enhancing LLM inference performance in resource‑constrained, practical deployment scenarios.

Furthermore, DSEI relies on a lightweight dynamic semantic extraction mechanism and introduces only approximately 0.6M additional parameters compared with SLLM. Despite this marginal parameter increase, it preserves the inherent efficiency advantages of semantic-level inference over token-by-token LLM inference in terms of both throughput and GPU memory overhead. Overall, these results demonstrate that DSEI provides a favorable trade-off between generation quality and inference efficiency.

\section{Conclusion}
In this paper, we propose the Dynamic Semantic Autoencoder (DSAE), which dynamically compresses sentences into segment-level latent representations. We integrate DSAE into Large Language Models (LLMs) via the Dynamic Semantic Extraction and Inference (DSEI) framework, enabling the LLM to conduct the entire inference process in the latent space. Our approach incorporates three key innovations: (1) employing a sentence segmentation mechanism to divide entire sentences into segments, thereby enhancing the model's semantic extraction capability for long sentences; (2) efficiently extracting dynamic semantics by assigning semantic weights to tokens through a lightweight contextual awareness mechanism; and (3) utilizing a gating mechanism to fuse dynamic and static semantics, which preserves global information while eliminating redundancy, thereby striking a balance between performance and efficiency. Experimental results demonstrate that, compared to static semantic extraction baselines, DSAE achieves a maximum improvement of 34\% in PPL, and DSEI reduces perplexity by 48\%. These results demonstrate that DSEI provides a practical trade-off between generation quality and deployment efficiency. It improves latent-space inference quality while preserving the computational advantages of compressed semantic representations.

\bibliographystyle{elsarticle-num} 
\bibliography{reference} 

\end{document}